\documentclass[10pt,usletter]{article}
\usepackage[totalwidth=420pt,totalheight=625pt]{geometry}

\newcommand{\cv}[1]{}
\newcommand{\av}[1]{#1}

\usepackage[T1]{fontenc}
\usepackage{amssymb}
\usepackage{url,hyperref}
\usepackage{xcolor}
\usepackage{enumitem}
\usepackage{inconsolata}

\usepackage[numbers,sort&compress]{natbib}
\usepackage{comment}
\usepackage{graphicx}
\usepackage{booktabs}
\excludecomment{CCSXML}
\newcommand{\ccsdesc}[2][]{}
\newcommand{\keywords}[1]{}

\usepackage{listings}
\lstdefinestyle{asp}{
  basicstyle=\ttfamily\normalsize,
  breaklines=false,
  columns=flexible,
  keepspaces=true,
  showstringspaces=false,
  xleftmargin=0pt,
  xrightmargin=0pt,
  commentstyle=\color{gray},
  morecomment=[l]{\%},
  literate={:-}{{\textbf{:-}}}2 {not}{{\textbf{not}}}3
}

\begin{document}

\title{ASP-Bench: From Natural Language to Logic Programs\thanks{To
    appear in the proceedings of the 2nd International Workshop on
    Neuro-Symbolic Software Engineering (NSE@ICSE 2026).
Supported by the Austrian Science Fund (FWF) projects
10.55776/P36688 and 10.55776/COE12.
}}

  \author{Stefan Szeider\\[4pt]
    \small Algorithms and Complexity Group\\[-3pt]
    \small TU Wien, Vienna, Austria\\[-3pt]
    \small \href{https://www.ac.tuwien.ac.at/people/szeider/}{www.ac.tuwien.ac.at/people/szeider/}
  }
  \date{}

\av{\maketitle}

\begin{abstract} %
  Automating the translation of natural-language specifications into
  logic programs is a challenging task that affects neurosymbolic
  engineering. We present ASP-Bench, a benchmark comprising 128
  natural language problem instances, 64 base problems with easy and
  hard variants. It evaluates systems that translate natural-language
  problems into Answer Set Programs (ASPs), a prominent form of
  logic programming. It provides systematic coverage of ASP features,
  including choice rules, aggregates, and optimization. Each problem
  includes reference validators that check whether solutions satisfy
  the problem specification.

  We characterize problems along seven largely independent reasoning
  aspects (optimization, temporal reasoning, default logic, resource
  allocation, recursion, spatial reasoning, and quantitative
  complexity), providing a multidimensional view of modeling
  difficulty.

  We test the benchmark using an agentic approach based on the
  ReAct (Reason and Act) framework, which achieves full saturation,
  demonstrating that feedback-driven iterative refinement with solver
  feedback provides a reliable and robust approach for modeling
  natural language in ASP. Our analysis across multiple agent runs
  enables us to gain insights into what determines a problem's
  modeling hardness.
\end{abstract}



\cv{\maketitle}

\section{Introduction}\av{\thispagestyle{empty}} %

Answer Set Programming (ASP) is a declarative programming approach
that captures many problems in knowledge representation and
reasoning~\cite{Gelfond1988}. Problems are encoded as logic programs
whose answer sets correspond to solutions.  In particular, ASP's
declarative nature supports modular development, compact problem
representation, and provably correct system design.  The paradigm
benefits from efficient general-purpose solvers such as
clasp~\cite{Gebser2012}, clingo~\cite{Gebser2019}, and
DLV~\cite{Leone2006}, which separate problem modeling from the solving
process.

Within ASP, problems are formulated as rules and constraints rather
than step-by-step algorithms.  Given a logic program, the solver
automatically computes solutions.  This is
similar to writing SQL queries rather than writing procedural code:
you describe what the solution must satisfy, not how to find it.

Translating natural language (NL) specifications into a logic
program remains a significant barrier to
the broader adoption of ASP, motivating research into automating this task.
Early approaches relied on Controlled Natural Languages
(CNLs), which ensure reliable translation by
restricting input to a formally defined grammar~\cite{Caruso2024}.
More recently, the advent of large language models (LLMs) has enabled
more flexible methods.  Prominent examples include multi-stage
pipelines that translate NL to a CNL
representation~\cite{Santana2024}, direct one-shot generation with
general-purpose LLMs~\cite{Ishay2023}, and fine-tuned domain-specific
models like LLASP~\cite{Coppolillo2024}.  However, experimental
evaluations reveal that one-shot generation is often inadequate;
Borroto et al.~\cite{Santana2024} found that ChatGPT-3.5 produced
flawed programs for 17 of 21 graph problems.  This suggests that a
more robust, agentic creation process is needed, in which agents use
solver feedback to refine and debug programs through multiple
reasoning-action cycles.

Other neurosymbolic systems, such as NeurASP~\cite{Yang2020}, assume a
human expert has already written a correct logic program.
Hence, neural networks provide only perceptual input (e.g.,
recognizing digits in images) that feeds into pre-existing
rules.  Differentiable solvers cast ASP computation as
graph neural networks for GPU parallelism~\cite{Skryagin2024} but
require the program itself to be manually encoded.

Recent work in NL2ASP translation has resulted in the NL-ASP-200
benchmark~\cite{Ishay2023}, comprising 200 problems with the best reported
accuracy of 58.5\%~\cite{Santana2024}. However, these benchmarks share critical limitations.
First, they evaluate using exact string matching.
Ishay et al.\ acknowledge that this approach rejects three out of four semantically correct programs that differ only syntactically from the reference solution.
Second, the problems are sampled opportunistically from online forums and textbooks, with limited systematic coverage of ASP language constructs.

Benchmarks for related formalisms---first-order logic~\cite{Han2024}, temporal logic~\cite{English2025}, propositional logic~\cite{Baek2025}, and constraint programming~\cite{Michailidis2025}---demonstrate that semantic verification using automated solvers is essential for reliable evaluation.

We present \textbf{ASP-Bench}, a benchmark designed to address these limitations.
ASP-Bench comprises 128 problem instances (64 base problems, each with easy and hard versions), providing systematic coverage of core ASP constructs, ranging from basic rules to choice rules, aggregates, and optimization.
Each problem is equipped with reference validators (``ground truth'') that enable semantic verification by checking whether solutions satisfy the problem specification, rather than requiring exact string matches.

To characterize the benchmark's diversity, we analyzed the 64 hard problem variants to identify recurring reasoning aspects that contribute to modeling difficulty.
Each problem incorporates 1 to 6 of these aspects, enabling analysis of which combinations of aspects make problems harder to model in ASP.

To demonstrate the benchmark and establish a baseline for future
comparisons, we apply an agentic approach based on the ReAct (Reason
and Act) framework~\cite{Yao2023}, which uses iterative refinement
with solver feedback.  Our system entails a single-prompt autonomous
LLM agent that performs all development and testing tasks without
following a fixed workflow or predefined pipeline.  Thus, in contrast
to one-shot generation methods~\cite{Ishay2023} or static composition
pipelines~\cite{Santana2024}, the agent dynamically decides its
actions based on solver feedback, using clingo error messages and
failed test cases to iteratively repair and improve ASP programs
across multiple reasoning-action cycles.  This approach achieves full
saturation on both easy and hard problems from ASP-Bench,
demonstrating the effectiveness of feedback-driven iterative
refinement for ASP program synthesis.  We also compare two MCP-based
approaches (one that shares the solver integration with our agent) for
interactive ASP modeling and find that general-purpose Python
execution outperforms specialized declarative interfaces.

\section{Example: Knights and Knaves} %

We provide an example of an NL problem and its translation into a logic program.
The example is based on Problem 03 (Knights and Knaves), one of the easy problem variants of ASP-Bench.
If a person is a knight, they always tell the truth; if they are a knave, they always lie.
Given that Alice says ``Bob is a knave'', Bob says ``Alice and Charlie are of the same type'', and Charlie says ``Alice is a knight'', the task is to determine each person's type.

The modeling is defined via first generating Python code using the clingo library:

\begin{lstlisting}[language=Python, basicstyle=\ttfamily\normalsize, stringstyle=\ttfamily]
from clingo import Control
ctl = Control()
ctl.add("base", [], """
person(alice; bob; charlie).
{ knight(X) : person(X) }.
knave(X) :- person(X), not knight(X).
...
""")
ctl.ground([("base", [])])
ctl.solve(on_model=extract_solution)
\end{lstlisting}

This Python program gives rise to the following logic program.

\begin{lstlisting}[style=asp]
person(alice; bob; charlie).
{ knight(X) : person(X) }.
knave(X) :- person(X), not knight(X).

:- knight(alice), knight(bob).    %
:- knave(alice), knave(bob).

:- knight(bob), knight(alice), knave(charlie).
:- knight(bob), knave(alice), knight(charlie).
:- knave(bob), knight(alice), knight(charlie).
:- knave(bob), knave(alice), knave(charlie).  %

:- knight(charlie), knave(alice). %
:- knave(charlie), knight(alice).
\end{lstlisting}

Key elements of this encoding are choice rules that generate candidates, completion axioms that define complementary predicates, and integrity constraints that enforce consistency.
Each statement gives rise to two constraints (truth if spoken by a knight, falsehood if by a knave).

\section{The ASP-Bench Benchmark} %

\subsection{Problem Construction Process}

We generated the 64 base problems in ASP-Bench with the help of a
coding agent. We proceeded through the following phases: brainstorming
problem ideas, drafting natural language problem descriptions,
creating ASP encodings, and testing solutions.  This process involved
iterative collaboration between the coding agent and human oversight.
This back-and-forth process produced 64 easy problem variants that
encompass diverse reasoning patterns, including logic puzzles, graph
problems, scheduling, optimization, and constraint satisfaction.
These include both completely new problems and variations of known
problems.  We used Claude Sonnet 4.5 as the primary LLM for the coding
agent Claude Code, with consultation support from Gemini Pro 2.5
(reasoning mode) via our consult7
tool\footnote{\url{https://pypi.org/project/consult7/}}, which offers
structured code analysis and reasoning capabilities by querying
several frontier models in parallel through the Model Context
Protocol (MCP)\footnote{\url{https://modelcontextprotocol.io/}}. This
provided valuable input for complex design decisions.

After establishing the 64 easy problems and their validators, we constructed hard variants by systematically increasing problem complexity.
The enhancement includes both quantitative (larger instances, more constraints) and qualitative (additional constraint types, multi-objective optimization, more intricate reasoning patterns) aspects.

For many hard problems, we used planted solutions: generating a known
feasible solution first, then constructing constraints around it.
This approach proved more reliable than unconstrained problem
generation because verifying that constraints admit solutions is
computationally easier than finding solutions to arbitrary constraint
systems.  Consequently, we can guarantee that hard problems are
well-posed and have solutions, a critical feature for benchmark
reliability and reproducibility.  We used multiple iterations of
agent-human interaction to ensure that problems remained solvable
within reasonable time bounds ($\leq$20 seconds for solver runtime).

\subsection{Semantic Verification}

ASP-Bench uses semantic verification rather than syntactic string matching.
Each problem has a \emph{ground truth} in the form of a reference validator: a Python script that checks whether a proposed solution satisfies the problem's specification.
The validator loads the JSON solution from standard input, verifies
that all problem constraints are satisfied, and outputs a validation
result in the format %

Each problem description explicitly specifies the required JSON fields
and their semantics, ensuring the solution format requirement is
unambiguous, and the probability of a correct solution in the wrong
format is very low.

The reference validator accepts any solution that correctly satisfies
the constraints. This is essential, as problems often have many valid
solutions, so simply comparing against a single fixed solution is not sufficient.

For optimization problems, we are also interested in the optimality of the solution.
The reference validator checks that the agent's solution matches the known optimum while satisfying all constraints.

We note that the validation is performed after the NL-to-ASP
construction is completed.  Hence, any approach tested on the
ASP-Bench must resort to its own testing and verification mechanisms,
as it has no access to the ground truth.

\subsection{Problem Characterization}\label{sec:characterization}

We analyze ASP-Bench problems along seven aspects representing
recurring reasoning patterns:

\begin{enumerate}
\item{OPT}: optimization problems using \texttt{minimize}/\texttt{maximize} statements;

\item {TEMP}: temporal or sequential reasoning requiring ordering constraints;

\item {DEFAULT}: problems with soft constraints or preferences;

\item {RESOURCE}: resource allocation with capacity limits;

\item {RECURSIVE}: reasoning requiring cycle detection or fixed-point computation;

\item {SPATIAL}: grid-based or neighborhood logic;

\item {QUANT}: high quantitative complexity (seven or more distinct constraints).
\end{enumerate}

These aspects are largely independent, allowing problems to incorporate multiple aspects simultaneously.

\begin{figure*}[tbh]
\centering
\includegraphics[width=\textwidth]{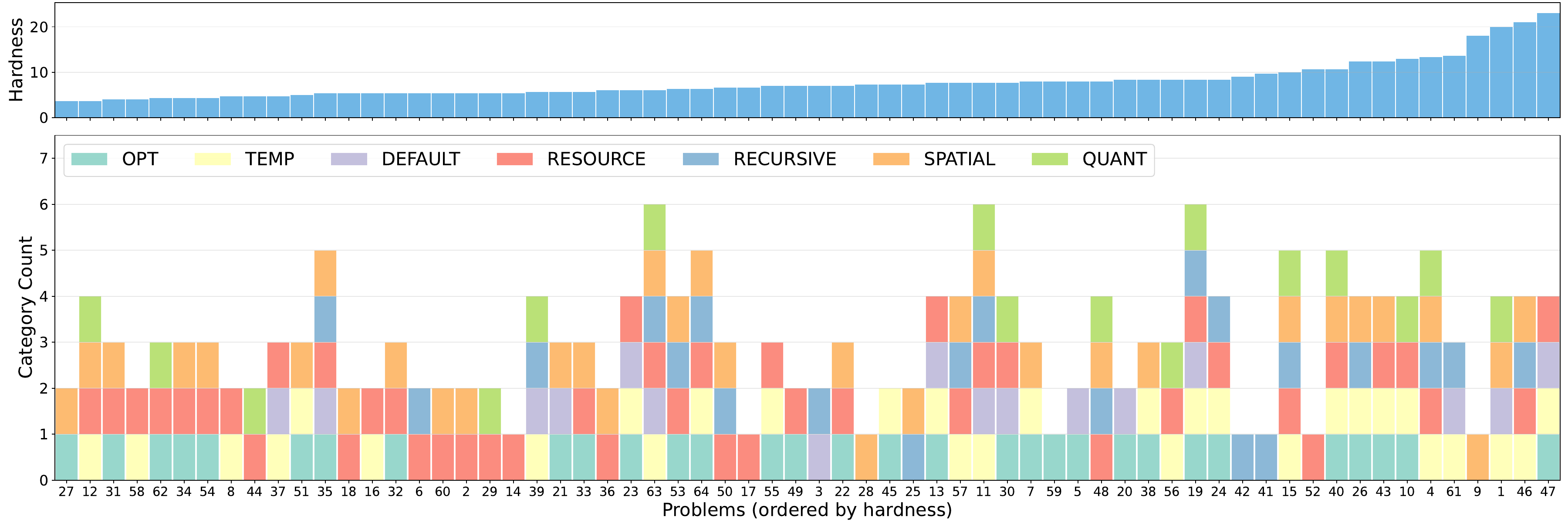}
\caption{Top: Average \texttt{python\_exec} calls per problem (hardness measure). Bottom: Reasoning aspects per problem. Problems are ordered by hardness (ascending).}
\label{fig:hardness}
\end{figure*}

The aspect taxonomy emerged from post-hoc analysis of the completed problem set, identifying common patterns that contribute to modeling difficulty in ASP.
Figure~\ref{fig:hardness} shows the distribution of the seven aspects over the 64 hard problem variants (ordered by ``hardness'' as will be described below).

\section{ASP Agent} %

The ASP-Agent is an adaptation of the general-purpose coding agent architecture that we introduced in the context of constraint modeling~\cite{Szeider2025b}.
The implementation is available as open-source software\footnote{\url{https://github.com/szeider/agentic-python-coder}}.
The agent builds on the ReAct (Reason and Act)
framework~\cite{Yao2023}, which facilitates an LLM to operate in an iterative cycle of reasoning, tool execution, and observation.
An important component of this architecture is a persistent IPython kernel that maintains state across multiple code executions, enabling the agent to build, test, and debug solutions incrementally.
This design cleanly separates domain-agnostic agent capabilities for code creation and execution from task-specific expertise, which is injected entirely via a project prompt rather than hard-coded into the agent's logic.

The agent uses two tools: \texttt{python\_exec} and \texttt{save\_code}. Repeated \texttt{python\_exec} calls drive the bulk of the iterative development. The number of \texttt{python\_exec} calls provides a  proxy for
problem hardness, reflecting the number of reasoning-action cycles the
agent  required to reach its  solution.
The \texttt{save\_code} tool is used just once at the end to signal
that the necoding is complete, and the final code is written to
disk. Running the code produces the solution in JSON format that the
reference validator checks.

The only two components of the agent that are specific for Answer Set
programming are (1)~a comprehensive project prompt
\texttt{clingo.md}\footnote{The prompt is available at
  \url{https://github.com/szeider/agentic-python-coder/blob/main/examples/clingo/clingo.md}}.
and (2) a parameter that loads the \emph{clingo} Python
library~\cite{Gebser2019}, which provides the necessary API for
modeling ASP problems and interacting with the clingo ASP solver.

The project prompt instructs the agent how to use the library.  It
further provides a detailed guide covering mandatory ASP syntax rules
(e.g., variable safety, aggregate placement), a proposed workflow,
common modeling patterns for scheduling and graph problems, and a
catalog of critical anti-patterns (e.g., incorrect cardinality in
choice rules, inefficient generate-and-filter logic).

The project prompt is organized into five main sections:

(1)~\emph{Mission Briefing \& Core Requirements}: establishing the
overall goal and non-negotiable constraints,

(2)~\emph{Critical Rules of Engagement}: detailing mandatory ASP
syntax, logical principles including the Closed World Assumption, and
the required 5-step workflow (plan, analyze \& model, implement, solve
\& extract, format \& verify),

(3)~\emph{Clingo API and Implementation Guide}: providing
Python-specific implementation guidance and the three-step action
modeling pattern for temporal problems,

(4)~\emph{Problem-Solving Pattern Library}: presenting reusable ASP patterns for common problem types, including assignment, graph, scheduling, and temporal reasoning problems, and

(5)~\emph{Debugging and Advanced Techniques}: covering performance
optimization strategies such as constraining choices early and
avoiding grounding explosions. This structured design provides the
agent with both foundational knowledge and practical guidance for
incrementally building and debugging ASP models.

We developed the ASP project prompt through an iterative refinement
process.  We tested an initial prompt containing basic API
instructions on a few easy problems from ASP-Bench.  We systematically
analyzed failures from the agent's execution logs, such as syntax
errors, unsafe variable declarations, and incorrect use of aggregates.
We updated the accordingly, adding explicit rules, best practices, and
concrete examples to address these pitfalls. However, we avoided overfitting to individual problems to enable generalization to unseen problems. We repeated this process over several iterations.
The resulting prompt was then evaluated on the full set of easy and hard problem variants. We observed a surprisingly fast learning rate and stopped the process after five iterations.

\section{Experimental Evaluation} %

\subsection{Experimental Setup}

We ran our ASP-Agent on each problem with Claude Sonnet 4.5 (September 29, 2025, version), accessed via OpenRouter, as the underlying language model, configured with temperature 0.0, max tokens 16384, and streaming enabled.
The hard problems were run with a 1200-second timeout for the agent-encoding process, while the easy problems were run with a 600-second timeout.
Each solver call  was limited to 20 seconds.
All runs completed well within the timeout.
We tested each hard problem on three independent runs to assess
consistency; we tested  easy problems in two runs.

The ASP-Bench benchmark set and all 128 problem descriptions, ground truth validators, and example solutions are available at Zenodo~\cite{ASPBench2025}.
We provide the data as an encrypted archive (password:
\texttt{qB5xjsoFtdSZWTm}) to prevent contamination of the training set in future experiments.

\subsection{Saturation on Easy and Hard Problems}

The ASP-Agent solved all instances correctly, including both runs on easy variants and all three runs on hard variants.

We report the \texttt{python\_exec} calls averaged across three runs
on each hard problem: 7.7 on average.  We observe  a substantial
variation in difficulty.  The easiest problem takes
3.7~\texttt{python\_exec} calls on average, the hardest 26.0.  The hardest
problems  were
problem~47 (DNA sequence assembly; 26.0), problem~46 (metroidvania
generation; 21.0), problem~01 (who is the killer; 20.0), problem~09
(nonogram solver; 17.2), and problem~61 (historical counterfactual;
13.7).  The easiest were problem~27 (queens domination; 3.7),
problem~12 (zebra puzzle; 3.7), problem~31 (network flow; 4.0), and
problem~58 (exam scheduling; 4.0).

The complete list of results is provided in Table~\ref{tab:results}, with execution counts and token usage averaged across multiple runs.

\begin{table*}[t]
\caption{ASP-Bench results: execution calls, and token usage
  ($\times 1000$\av{, I=in, O=out}) averaged across runs (2 for easy, 3 for hard).}
\label{tab:results}
\centering

\cv{\setlength{\tabcolsep}{3pt}
  \renewcommand{\arraystretch}{0.85}}

\av{\setlength{\tabcolsep}{1.7pt}
  \renewcommand{\arraystretch}{0.75}}

\begin{tabular}{@{}rl rrr rrr c rl rrr rrr@{}}
\toprule
 & & \multicolumn{3}{c}{Easy} & \multicolumn{3}{c}{Hard} & & & &
                                                                 \multicolumn{3}{c}{Easy} & \multicolumn{3}{c}{Hard} \\ 
\cmidrule(lr){3-5} \cmidrule(r){6-8} \cmidrule(l){12-14}
  \cmidrule(lr){15-17}
\cv{  \# & Problem & Ex & In & Out & Ex & In & Out &\hspace{6mm} & \# & Problem & Ex & In & Out & Ex & In & Out \\}
\av{  \# & Problem & Ex & I & O & Ex & I & O &\hspace{2mm} & \# & Problem & Ex & I & O & Ex & I & O \\}

  \midrule
01 & Who Is Killer       &  4.5 & 154 &  4 & 20.0 & 622 & 23 & & 33 & Independent Set     &  3.5 & 126 &  3 &  5.7 & 154 &  7 \\
02 & Wtd Graph Color     &  3.0 & 119 &  3 &  5.3 & 155 &  6 & & 34 & Dominating Set      &  4.0 & 136 &  3 &  4.3 & 116 &  5 \\
03 & Knights Knaves      &  2.0 &  96 &  2 &  7.4 & 202 & 10 & & 35 & Feedback Vtx Set    &  5.5 & 174 &  4 &  5.3 & 141 &  6 \\
04 & Blocks World        &  4.5 & 163 &  4 & 10.8 & 314 & 11 & & 36 & Latin Square        &  2.0 &  99 &  3 &  6.0 & 149 &  5 \\
05 & Circuit Diagnosis   & 13.5 & 588 & 23 &  9.2 & 355 & 18 & & 37 & Car Sequencing      &  3.5 & 133 &  4 &  4.7 & 126 &  5 \\
06 & Hospital Matching   &  3.0 & 123 &  5 &  5.0 & 270 &  7 & & 38 & Protein Structure   &  4.5 & 165 &  5 &  8.3 & 208 &  7 \\
07 & Hamilton P\cv{ath} Wtd   &  4.0 & 141 &  3 &  8.0 & 196 &  7 & & 39 & Byzantine Generals  &  4.0 & 136 &  3 &  5.7 & 181 &  9 \\
08 & Meeting Sched       &  4.5 & 175 &  5 &  4.7 & 130 &  7 & & 40 & Warehouse \cv{Network}   &  3.0 & 120 &  4 &  9.0 & 402 & 19 \\
09 & Nonogram Solver     &  3.5 & 152 &  7 & 17.2 & 626 & 20 & & 41 & Argumentation \cv{Fwk}   &  5.0 & 155 &  3 &  9.7 & 279 &  8 \\
10 & Resource Alloc I    &  4.5 & 160 &  4 & 10.0 & 324 & 14 & & 42 & Gene Reg Network    &  2.0 &  97 &  3 &  4.3 & 137 &  5 \\
11 & Tournament Rank     &  3.5 & 129 &  3 &  7.7 & 209 &  8 & & 43 & Quantum Circuit     &  4.5 & 169 &  5 & 12.3 & 377 & 19 \\
12 & Zebra Puzzle        &  4.0 & 160 &  7 &  3.7 & 112 &  6 & & 44 & Nontransitive Dice  &  3.0 & 137 &  4 &  4.7 & 135 &  5 \\
13 & Job Shop Sched      &  3.5 & 137 &  5 &  7.7 & 220 & 11 & & 45 & Prisoners Dilemma   &  7.5 & 223 &  6 &  7.3 & 211 &  7 \\
14 & Cryptarithmetic     &  3.0 & 118 &  3 &  5.3 & 141 &  5 & & 46 & Metroidvania Gen    &  7.0 & 232 &  7 & 21.0 & 929 & 30 \\
15 & Traveling Tourn     & 10.5 & 339 &  8 & 10.0 & 321 & 11 & & 47 & DNA Seq\cv{uence} Asm    &  6.0 & 194 &  5 & 26.0 & 922 & 23 \\
16 & Nurse Rostering     & 19.5 & 724 & 17 &  5.3 & 128 &  4 & & 48 & Crossword           & 12.0 & 360 & 10 &  8.0 & 229 &  9 \\
17 & Bin Packing         &  5.0 & 163 &  4 &  6.7 & 181 &  7 & & 49 & Auction \cv{Mechanism}   &  2.5 & 110 &  4 &  7.0 & 189 &  8 \\
18 & Magic Square        &  5.0 & 161 &  3 &  5.3 & 136 &  6 & & 50 & Cellular Automata   &  6.0 & 192 &  6 &  6.7 & 188 &  6 \\
19 & Course Timetable    &  3.0 & 123 &  4 &  8.3 & 234 & 11 & & 51 & Ricochet Robots     &  6.5 & 223 &  9 &  5.0 & 159 &  9 \\
20 & Set Cover           &  4.0 & 139 &  3 &  8.3 & 213 &  7 & & 52 & Nim Game            &  5.0 & 164 &  5 & 10.7 & 291 & 11 \\
21 & Wtd Vertex Cover    &  3.0 & 119 &  3 &  3.3 & 114 &  4 & & 53 & Steiner Tree        &  4.0 & 144 &  5 &  6.3 & 176 & 10 \\
22 & Clique Finding      &  4.5 & 148 &  3 &  7.0 & 168 &  6 & & 54 & Graph Partition     &  2.5 & 110 &  3 &  4.3 & 118 &  5 \\
23 & Resource Alloc II   &  3.0 & 121 &  3 &  6.0 & 160 &  8 & & 55 & Recipe Planning     &  6.5 & 213 &  7 &  7.0 & 195 &  9 \\
24 & Workflow Optim      &  5.5 & 205 &  5 &  8.3 & 201 &  7 & & 56 & Music Composition   &  4.5 & 150 &  4 &  8.3 & 225 &  9 \\
25 & Sudoku Mines        &  6.0 & 189 &  5 &  7.7 & 257 &  9 & & 57 & Escape Room         &  5.0 & 161 &  4 &  7.7 & 233 & 11 \\
26 & Tower Of Hanoi      &  6.5 & 213 &  5 & 12.3 & 368 & 12 & & 58 & Exam Scheduling     &  5.5 & 178 &  5 &  4.0 & 110 &  5 \\
27 & Queens Dom\cv{ination}   &  4.5 & 152 &  3 &  3.7 &  95 &  3 & & 59 & Strategic Voting    & 11.0 & 388 & 12 &  8.0 & 228 & 10 \\
28 & Graph Iso\cv{morphism}   &  5.0 & 166 &  4 &  7.3 & 218 &  9 & & 60 & Ecosystem Balance   &  6.0 & 199 &  7 &  5.3 & 133 &  4 \\
29 & Logic Grid Puzzle   &  3.5 & 130 &  4 &  5.3 & 154 &  8 & & 61 & Historical Cfact    &  4.0 & 143 &  4 & 13.7 & 374 & 12 \\
30 & Team Formation      &  4.5 & 149 &  4 &  7.7 & 211 &  9 & & 62 & Drug Interaction    &  8.5 & 302 & 12 &  4.3 & 129 &  7 \\
31 & Network Flow        &  4.5 & 157 &  5 &  4.0 & 120 &  6 & & 63 & Dungeon Gen         & 11.0 & 396 & 18 &  6.0 & 205 & 13 \\
32 & Frequency Assign    &  2.5 & 125 &  4 &  5.3 & 138 &  5 & & 64 & Social Net Infl     &  7.0 & 225 &  7 &  6.3 & 176 &  8 \\
\bottomrule
\end{tabular}
\end{table*}

\subsection{Problem Hardness and Reasoning Aspects}\label{sec:aspects}

To characterize what makes problems difficult for the agent, we
investigated the reasoning aspects of each problem as defined in
Section~\ref{sec:characterization}.
The results are shown in Figure~\ref{fig:hardness}.

Surprisingly, problem hardness (measured by the number of
\texttt{python\_exec} calls) shows no significant correlation with the
number of reasoning aspects.  The execution counts range between
 5.9 for two-aspect problems (the easiest group) to 10.8 for
four-aspect problems (the hardest group on average), with
single-aspect problems averaging 9.3 executions.  This suggests that
some problem domains are considerably harder for our agent than
others, independent of the number of reasoning aspects; difficulty
arises from the depth and complexity of reasoning within specific
domains, rather than the breadth across multiple reasoning aspects alone.

Let us look at an example to illustrate this pattern.  Problem~9
(nonogram solver) involves only the SPATIAL aspect.  It is one of the
hardest problems, with an average of 17.2 executions, due to complex
grid-constraint reasoning.  On the other hand, Problem~27
(queens domination) combines SPATIAL and OPT aspects but is among the
easiest, with 3.7 executions, as the constraint structure admits
straightforward modeling.  Problem~47 (DNA sequence assembly),
which combines four aspects (OPT, TEMP, DEFAULT, and RESOURCE), is the
hardest, with 26.0 executions.  We believe its difficulty is due
to intricate overlap constraints rather than aspect breadth.

\subsection{Agent Activity Patterns}

To understand how the agent allocates effort as problems become
harder, we performed a bottom-up analysis of agent transcripts from
all $3\times 64$ runs on the hard problems, categorizing each execution attempt into eight distinct activity types:

\emph{Data Setup}: problem scoping and data initialization;
\emph{Initial Model}: first ASP model construction;
\emph{Rule Addition}: incremental constraint addition;
\emph{Logic Verification}: targeted testing of specific rules and constraints;
\emph{Error Correction}: debugging and fixing errors;
\emph{Full Solve}: complete solution attempts;
\emph{Logic Refinement}: modifying existing ASP rules to correct errors;
\emph{Solution Formatting}: finalizing output format and presentation.

\begin{figure}[t]
\centering
\begin{minipage}[t]{0.49\columnwidth}
\centering
\includegraphics[width=\linewidth]{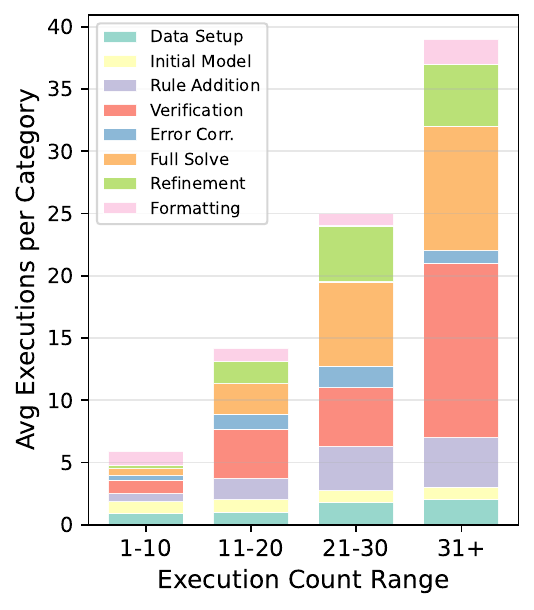}
\end{minipage}%
\hfill
\begin{minipage}[t]{0.49\columnwidth}
\centering
\includegraphics[width=\linewidth]{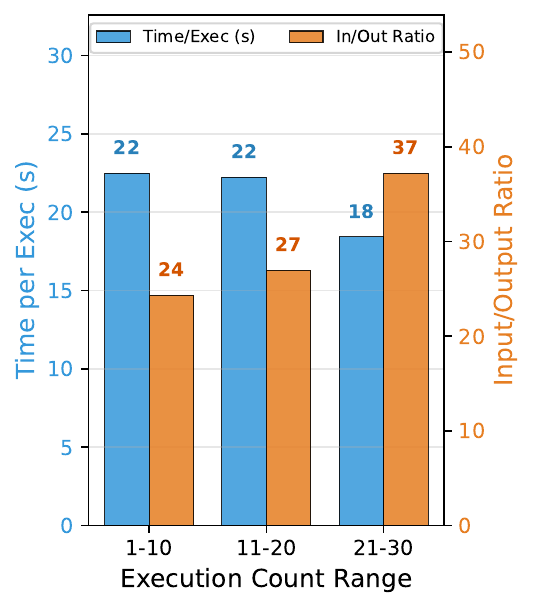}
\end{minipage}
\caption{Left: Distribution of agent activity types across difficulty ranges (1--10, 11--20, 21--30, 31+ executions). Right: Agent behavior metrics showing time per execution (blue, left axis) and input/output token ratio (orange, right axis).}
\label{fig:activity}
\end{figure}

We split runs into four difficulty ranges based on total execution
count: $\leq 10$ executions (easier problems), 11--20 executions (moderate
difficulty), 21--30 executions (challenging problems), and $\geq 31$  executions (hardest problems).
Figure~\ref{fig:activity} shows the analysis of activity distribution
across these ranges. It reveals how the agent's behavior adapts to the problem's complexity.
For easier problems (1--10 executions), we observe relatively even effort allocation across activities, with quick progression from initial model through verification to solution.
As problems become harder (11--20 and 21--30 execution calls), logic verification accounts for a larger share of the activity distribution.
This activity consumes an increasing proportion of total execution calls.
This highlights that the harder problems require a careful incremental approach. The agent tests individual constraints and rules before attempting complete solutions.
For the hardest problems (31+ executions, represented by only two runs: DNA sequence assembly and nonogram solver), the number of full solve attempts is much higher as the agent iteratively refines complete models through multiple cycles.
Error correction is below 10\% of activities, even for the hardest problems.
This suggests that the iterative refinement approach provides feedback that prevents errors rather than requiring extensive post-hoc debugging.

The right panel of Figure~\ref{fig:activity} shows two additional patterns: time per execution decreases from 22 seconds for easy problems to 15 seconds for the hardest, while the input/output token ratio increases from 24:1 to 45:1.
Harder problems involve shorter, more focused execution cycles where the agent reads proportionally more context before generating each response.

\section{Anatomy of Agentic Modeling} %

To understand how the ASP-Agent applies the project prompt in
practice, let us examine Problem 26 (Tower of Hanoi) in detail.
This problem demonstrates the agent's flexibility in problem-solving strategies, showing dramatically different execution patterns across runs: the shortest run completed in 6~\texttt{python\_exec} calls, while the longest required 20~calls, a more than 3-fold variance that reveals fundamentally different approaches to the same problem.

\subsection{Problem Complexity: The Pilgrim's Journey} %

The problem is a variant of the classic 4-disk, 4-peg Tower of Hanoi
Problem, with the additional requirement that every disk must visit
both intermediate pegs B and C during its journey from source peg A to
destination peg~D.  This ``Pilgrim's Journey'' constraint rules out
standard Tower of Hanoi algorithms.  This transforms the problem from
algorithmic execution into a complex state-space search.  The expected
optimal solution requires 19~moves, providing a challenging search depth
bound for temporal planning.  The formulation operates with the
following parameters: state representation (disk positions across 4
pegs at each time step), actions (valid move operations respecting
standard Hanoi rules), temporal constraints (disk trajectories must
include mandatory visits to pegs B and C), and optimization (minimize
total moves while satisfying all constraints).

\subsection{The Expert Strategy (6 Calls)} %

In the shortest run, our agent demonstrated expert-like behavior: it
immediately recognized the problem structure as a temporal planning
problem. It formulated a complete, accurate ASP model with minimal iteration.

\textbf{Calls 1--2: Problem analysis and data setup.}
The agent read the problem description, extracted problem parameters (4~disks, 4~pegs, 19~move limit), and defined state predicates and action schema.

\textbf{Calls 3--4: Core ASP model construction.}
The agent built a comprehensive model holistically, defining: (1)~state fluent \texttt{on(Disk, Peg, Time)} representing disk positions, (2)~action \texttt{move(Disk, FromPeg, ToPeg, Time)} with preconditions and effects, (3)~frame axioms for persistence of disk positions between moves, and (4)~standard Hanoi constraints ensuring no disk sits above a smaller disk.

\textbf{Call 5: Pilgrim's Journey constraints.}
It further added temporal trajectory constraints to ensure that each disk visits both intermediate pegs, using a \texttt{visited/2} predicate to track which pegs each disk has visited during the sequence.

\textbf{Call 6: Verification and output.}
The agent executed the clingo solver (found a solution in less than
1~second), verified that the move sequence satisfies all constraints, and formatted JSON output with the complete move sequence.

This shows that the linear progression (analyze, model, constrain,
solve) emerges when the agent is operating at peak efficiency.
Further debugging was not necessary because the model was correct on
the first attempt. Here is the agent's output.

\begin{lstlisting}[style=asp]
top(D,P,T) :- on_peg(D,P,T), not has_smaller(D,P,T).
has_smaller(D,P,T) :- on_peg(D,P,T), on_peg(D2,P,T),
    D2 < D.
1 { move(D,P1,P2,T) : top(D,P1,T), peg(P2),
                      P1!=P2 } 1 :- time(T).
on_peg(D,P2,T+1) :- move(D,_,P2,T).
on_peg(D,P,T+1) :- on_peg(D,P,T), not move(D,P,_,T).
visited(D,P) :- move(D,_,P,_).
:- disk(D), not visited(D,b).
:- disk(D), not visited(D,c).
\end{lstlisting}

\subsection{The Debugging Strategy (20 Calls)} %

The longest run reveals a remarkably different approach, demonstrating sophisticated debugging methodology when initial attempts failed.
The agent's own summary explicitly notes: ``Initial attempts with 19~steps failed due to the solver struggling with the large search space. Testing with smaller instances (1--3~disks) helped verify the model correctness.''

\textbf{Calls 1--8: Initial model attempts and challenges in the
  search space.}  The agent built an initial ASP model, but the solver
failed to find solutions within a reasonable time.  The model was
either overly complex or contained inefficiencies that prevented
practical solving.  Instead of simply tweaking the same model
repeatedly, the agent pivoted to a diagnostic strategy.

\textbf{Calls 9--14: Simplification for verification.}
Recognizing the performance issues, the agent implemented a clever meta-cognitive strategy: simplifying the problem from 4~disks to 1--3~disks, reducing the time horizon to match the smaller problem scale, testing core ASP logic on these tractable instances, and verifying that the rules correctly encoded Hanoi constraints and state transitions.
This teach-yourself-on-an-easier-version approach is a sophisticated strategy: the agent generated a hypothesis (the logic is correct; scaling is the issue) and systematically tested it.
The smaller instances solved successfully in less than 0.2~seconds, confirming the core modeling approach was sound.

\textbf{Calls 15--18: Scaling and optimization.}
With confidence in its logic, the agent scaled back to 4~disks and refined the model by tightening bounds on intermediate variables, adding symmetry-breaking constraints, reformulating some rules for more efficient grounding, and re-testing with incremental time horizons (12, 15, 19~moves).

\textbf{Calls 19--20: Final solution and verification.}
The optimized model was then solved within time limits.
The agent verified the solution trajectory, confirming all 4~disks visited both B and C, and formatted the output.

This run demonstrates the agent's resilience and strategic flexibility.
Instead of simply retrying when faced with failure, the agent diagnosed, simplified, verified, and scaled.
This result is in line with the findings on activity patterns we presented in Figure~\ref{fig:activity}: longer runs result in more verification cycles, incremental testing, and strategic refinement before reaching the solution.
The following listing shows the final ASP encoding with the refined approach using explicit preconditions.

\begin{lstlisting}[style=asp]
top(D,P,S) :- on(D,P,S), not blocked(D,P,S).
blocked(D,P,S) :- on(D,P,S), on(D2,P,S), D2 < D.
1 { move(D,P1,P2,S) : disk(D), peg(P1), peg(P2),
                      P1!=P2 } 1 :- step(S).
:- move(D,P1,_,S), not top(D,P1,S-1).
:- move(D,_,P2,S), top(D2,P2,S-1), D > D2.
on(D,P2,S) :- move(D,_,P2,S).
on(D,P,S) :- on(D,P,S-1), not move(D,P,_,S).
visited(D,P) :- move(D,_,P,_).
:- disk(D), not visited(D,b).
:- disk(D), not visited(D,c).
\end{lstlisting}

\subsection{Contrasting Strategies} %

The high variance in execution calls translates to dramatic differences in resource consumption.
The expert run required 101~seconds using 140k input tokens and 4.8k output tokens, while the debugging run required 574~seconds with 690k input tokens and 19.7k output tokens---approximately 5~times more resources.
Evidently, both runs have the same optimal solution with 19~moves.

\section{MCP Integration} %

Our autonomous modeling system can also be run by swapping our custom
ReAct agent for another agentic system, such as a coding agent or a
chatbot application, and connecting to our solver interface via the Model
Context Protocol (MCP).  This protocol provides a standardized
open-source interface for connecting language models with external tools
and data sources~\cite{MCP2024,Hou2025MCP}.  Introduced by Anthropic in
late 2024 and now hosted by the Linux Foundation, MCP has become a de
facto standard for agent-tool interaction, with adoption by major AI
platforms and integration into development environments.

\sloppypar We compare two MCP-based approaches for ASP modeling.

The first, \textbf{ipython-mcp}, is what we get when we replace the
ReAct agent in our system with an external agentic system. This MCP
server exposes a persistent IPython kernel with MCP tools for code
execution (\texttt{python\_exec}), state management
(\texttt{python\_reset}), and session control. Domain adaptation is
achieved through the project prompt (\texttt{clingo.md}), which
mirrors our standalone agent architecture with full Python control
flow.

The second, \textbf{mcp-solver-asp}~\cite{Szeider2025a},
provides specialized declarative model editing through item-based
tools (\texttt{add\_item}, \texttt{replace\_item},
\texttt{delete\_item}, \texttt{get\_model}, \texttt{solve\_model}),
guided by an instructions prompt included in the MCP-Solver
distribution. This enforces a purely declarative workflow where the
agent manipulates individual ASP rules rather than procedural code.

\begin{table}[t]
\centering
\caption{MCP approach comparison on 10 hardest ASP-Bench problems (3 runs each). Column \textbf{p} shows successful runs out of 3; tool calls and time are averaged.}
\label{tab:mcp}
\av{\smallskip}
\setlength{\tabcolsep}{4pt}
\begin{tabular}{@{}lrrrrrr@{}}
\toprule
& \multicolumn{3}{c}{\textbf{ipython-mcp}} & \multicolumn{3}{c}{\textbf{mcp-solver-asp}} \\
\cmidrule(r){2-4} \cmidrule(l){5-7}
\textbf{Problem} & \textbf{p} & \textbf{calls} & \textbf{time} & \textbf{p} & \textbf{calls} & \textbf{time} \\
\midrule
01 Who Is Killer      & 3 & 25.3 & 277 & 3 & 61.0 & 290 \\
04 Blocks World       & 3 &  9.3 & 170 & 2 & 34.7 & 276 \\
09 Nonogram           & 3 & 20.0 & 207 & 3 & 51.0 & 276 \\
10 Resource Alloc     & 3 & 11.0 & 166 & 2 & 53.3 & 327 \\
26 Tower of Hanoi     & 3 & 19.0 & 300 & 3 & 44.7 & 338 \\
43 Quantum Circuit    & 3 & 14.7 & 254 & 2 & 44.0 & 292 \\
46 Metroidvania       & 3 & 18.0 & 248 & 3 & 49.0 & 330 \\
47 DNA Sequence       & 3 & 18.7 & 278 & 3 & 51.7 & 344 \\
52 Nim Game           & 3 &  9.7 & 137 & 3 & 35.0 & 252 \\
61 Historical Cfact   & 3 & 14.0 & 121 & 3 & 48.0 & 249 \\
\midrule
\textbf{Total/Average} & \textbf{30} & \textbf{16.0} & \textbf{216} & \textbf{27} & \textbf{47.2} & \textbf{297} \\
\bottomrule
\end{tabular}
\end{table}

\paragraph{Experiment} We selected the 10 hardest problems from ASP-Bench based on the average execution counts in Table~\ref{tab:results}.
We ran each of the two MCP systems three times per problem, for a total
of 60 runs.
For this comparison, we used Claude Sonnet 4.5 for both MCP systems
with identical timeout settings (1200 seconds).
Table~\ref{tab:mcp} shows the results: ipython-mcp achieved 100\% accuracy (30/30 runs), while mcp-solver-asp achieved 90\% (27/30), with failures on blocks world, resource allocation, and quantum circuit.
We also observe that ipython-mcp uses significantly fewer tool calls (16.0 vs.\ 47.2 on average) and less time (216 seconds vs.\ 297 seconds) than mcp-solver-asp.

\paragraph{Discussion} This performance difference can be explained by
architectural trade-offs.  The ipython-mcp approach readily supports
full Python expressiveness, as the agent can construct ASP programs as
strings, use loops for repetitive constraints, implement custom
parsing and solution extraction, and leverage Python's debugging
capabilities.  The mcp-solver-asp approach enforces item-by-item model
construction, requiring more tool calls and limiting procedural
abstractions.  However, mcp-solver-asp's specialized interface offers
potential advantages not captured in this comparison: item-based
editing provides finer-grained undo/redo capabilities, makes model
state explicitly inspectable, and prevents certain error classes,
such as string-quoting issues in embedded ASP. It also includes code
analysis that prevents possibly harmful code from operating, whereas
ipython-mcp has no fixed guardrails.  In educational contexts or for
interactive exploration, the declarative interface may be more
intuitive.

\enlargethispage*{5mm}
\section{Conclusion} %

We have presented ASP-Bench, a benchmark for evaluating automated
translation from natural language to Answer Set Programming,
comprising 128 problem instances (64 base problems with easy and hard
variants). The benchmark provides systematic coverage of core ASP
constructs and supports semantic verification that checks whether
solutions satisfy the problem specifications.

With an agentic approach based on the ReAct framework, we achieved
full saturation on both easy and hard problem sets across multiple
runs.  This highlights the effectiveness of iterative refinement with
solver feedback.  This approach is not restricted to fixed pipelines.

Although full saturation was achieved, we believe that the benchmark
provides a valuable basis for neurosymbolic AI research. It provides
researchers with a rigorous testbed featuring diverse reasoning
patterns, semantic validation, and diagnostic capabilities for
identifying system failures on specific ASP features. In particular,
the benchmark can be used to compare approaches across various
efficiency metrics, including token consumption, execution cycles, and
wall-clock time.

The next generation of NL-to-ASP systems should aim to reduce the
number of reasoning-action cycles and token consumption while
maintaining correctness and accuracy.  Such improvements could be
achieved through automated prompt optimization with systems like
DSPy~\cite{Khattab2024}, which could also enable ablation studies to
identify which sections of the project prompt contribute most to agent
performance.  The monolithic project prompt could also benefit from
progressive disclosure, where the agent selectively loads relevant
sections based on problem characteristics, reducing token consumption
while maintaining access to specialized guidance.

Additionally, ASP-Bench can be used to evaluate smaller open-weight
models, prompt optimization systems, and alternative agent
architectures where correctness may not reach saturation.
Another direction would be to achieve comparable coverage with locally
deployable, open-weight LLMs.  This makes automated ASP programming
accessible without dependence on commercial API services.

The design of hard benchmark problems, together with reference
validators, is a nontrivial task, a fact we learned when designing the
hard problems of ASP-bench. A challenging objective for future work is
the design of a new category of ``very hard'' problems to be included
in an updated ASP-Bench.

Beyond generating correct ASP encodings, neurosymbolic approaches can
optimize encodings for solver performance by exploring alternative
constraint formulations and identifying more efficient problem
representations; we hope that, also in this direction, ASP-Bench provides a solid
basis.


\end{document}